\documentclass{article} 
    \makeatletter
\def\@fnsymbol#1{\ensuremath{\ifcase#1\or \dagger\or \ddagger\or
   \mathsection\or \mathparagraph\or \|\or **\or \dagger\dagger
   \or \ddagger\ddagger \else\@ctrerr\fi}}
    \makeatother
\usepackage{iclr2024_conference,times}

\usepackage{amsmath,amsfonts,bm}

\def\eqref#1{equation~\ref{#1}}

\def\1{\bm{1}}

\DeclareMathAlphabet{\mathsfit}{\encodingdefault}{\sfdefault}{m}{sl}
\SetMathAlphabet{\mathsfit}{bold}{\encodingdefault}{\sfdefault}{bx}{n}

\usepackage{hyperref}
\usepackage{url}
\usepackage{graphicx}
\usepackage{tabularx}
\usepackage{booktabs}
\usepackage{makecell}
\usepackage{multirow}
\usepackage{mathtools}
\iclrfinalcopy

\title{URL-BERT: Training Webpage Representation via Social Media Engagements}

\author{Ayesha Qamar\thanks{this work was done when the author was working at Twitter.} \\
Department of Computer Science\\
Texas A\&M University\\
\texttt{ayesha@tamu.edu} \\
\And
Chetan Verma\footnotemark[1] \\
VerSe Innovation Labs \\
\texttt{chetan.verma@verse.in} \\
\And
Ahmed El-Kishky \& Sumit Binnani \\
Twitter Inc \\
\texttt{\{aelkishky,sbinnani\}@twitter.com} \\
\AND
Sneha Mehta\footnotemark[1] \\
Independent \\
\texttt{smehta921@gmail.com}
\AND
Taylor Berg-Kirkpatrick\footnotemark[1] \\
UC San Diego \\
\texttt{tberg@ucsd.edu}
}

\begin{document}
\maketitle
\begin{abstract}
Understanding and representing webpages is crucial to online social networks where users may share and engage with URLs. Common language model (LM) encoders such as BERT can be used to understand and represent the textual content of webpages. However, these representations may not model thematic information of web domains and URLs or accurately capture their appeal to social media users. In this work, we introduce a new pre-training objective that can be used to adapt LMs to understand URLs and webpages. Our proposed framework consists of two steps: (1) scalable graph embeddings to learn shallow representations of URLs based on user engagement on social media and (2) a contrastive objective that aligns LM representations with the aforementioned graph-based representation. We apply our framework to the multilingual version of BERT to obtain the model URL-BERT. We experimentally demonstrate that our continued pre-training approach improves webpage understanding on a variety of tasks and Twitter internal and external benchmarks.
\end{abstract}

\section{Introduction}

On many social networks, users can share web content with other users by posting URL addresses to webpages. As such, understanding the semantic content of these shared webpages is crucial not only for health and safety initiatives, but also to topically categorize and recommend these webpages to social media users. 

Large pre-trained language models~\citep{devlin2018bert,zhang2022twhin} based on the Transformer architecture~\citep{vaswani2017attention} have become a standard tool for content understanding. These models are trained on general-domain corpora and can be used to represent a variety of content including webpages. However, despite the versatility of pre-trained language models such as BERT~\citep{devlin2018bert}, these models fail to capture thematic web-domain and URL-based signals. Additionally, text self-supervision pre-training such as masked language modeling fails to capture web-domain appeal to users. A LM would encode two webpages closer together if they share similar content but doing so disregards user preferences. For example, two news articles reporting on the same event can vary in their reporting style and therefore each maybe of interest to a different audience.

A valuable and distinctive feature of social media is explicit user engagement with shared content such as URLs. For example, on Twitter, Users may ``Favorite'', ``Reply'', ``Share'' or ``Retweet'' Tweets containing URLs. With the assumption that Users who are interested in similar content largely engage with similar webpages, this engagement can be an invaluable signal to webpage understanding.

In this work, we utilize relational user to URL engagements to continue pre-training for BERT. The key idea of our method is to first construct a User to URL engagement graph and perform scalable graph embedding~\citep{el2022twhin, el2022knn} to learn URL representations. We then continue pre-training to adapt the pre-trained BERT model to webpages. This continued pre-training of BERT takes the URL and content of a webpage and attempts to align the generated contextualized embedding with the aforementioned trained engagement-based URL embeddings. Many of the tasks that require webpage understanding cannot benefit directly from graph engagement-based representations since those methods are transductive and cannot be directly used to derive representations for content not seen during training. It is not feasible to retrain graph-based representations every time new data comes in. This is particularly prohibitive for applications such as spam filtering where the newer content is of high importance. We train engagement-based URL embeddings for 30 million URLs using 20 billion User-to-URL engagements and utilize these URL embeddings to train our URL-BERT model.

Our proposed approach facilitates the LM to better represent a webpage using not only the semantic content but also supervision provided by user engagement. As a result of this, when such an encoder is used on a downstream task,  URL-BERT needs only a few examples to show improvements as compared to the LM that it is based on. To demonstrate this, we evaluate our trained URL-BERT model on few-shot setting for webpage topic classification, Tweet hashtag prediction, and user engagement tasks and demonstrate that the URL-BERT model outperforms baseline BERT representations. 

In particular, our main contributions are threefold 
\begin{enumerate}
\item To instill knowledge learned through graph-based representations into LMs, we present a contrastive learning-based pre-training objective.
\item We show an LM pre-trained in such a manner can implicitly capture user-content engagement and in turn produce better representations when only given the content.
\item We show the effectiveness of this approach on several downstream tasks.
\end{enumerate}

\section{Related Works}
\textbf{Pre-trained Language Models:}
Since their introduction, pre-trained language models (PLM) ~\citep{peters2018deep, devlin2018bert} have enjoyed success as building blocks for many natural language processing tasks. A large number of approaches have been developed to train content representations using text-based self-supervision. Most prominently of these is BERT ~\citep{devlin2018bert} which is trained using a masked language model (MLM) objective and next sentence predictions. RoBERTa ~\cite{liu2019roberta} provided a rigorous hyper-parameter optimization and deduced that MLM was sufficient as a sole objective. Later PLM variants ~\citep{raffel2020exploring, lan2019albert, sanh2019distilbert, yang2019xlnet} applied similar text self-supervision approaches for pre-training. Continual pre-training involves initializing a LM with trained weights and then continuing to train on either in-domain data or a modified objective~\citep{kalyan2021ammus} this method is mostly used to adopt a PLM to a specific domain~\citep{lee2020biobert,wu2020tod,barbieri2022xlm}. The advantage of doing continual pre-training is not having to train a LM from scratch, which can be computationally expensive. Our proposed framework continues pre-training with an engagement objective after the initial MLM pre-training.

\noindent
\textbf{Webpage Representation:}
Another line of work focuses particularly on utilizing structural information such as HTML to represent webpages~\citep{deng2022dom, li2021markuplm, reis2004automatic}. Some works incorporate rich features extracted from webpages such as text \citep{buber2019web, abidin2016algorithm}, images \citep{manugunta2022deep, lopez2018deep, lopez2019visual}, while others combine both textual and visual features \citep{liparas2014news, kovacevic2002recognition, fersini2008enhancing}. Lastly, there are works that focus solely on the URL link itself, ignoring the textual content ~\citep{baykan2009purely, baykan2011comprehensive, hernandez2014cala, abdallah2014url}. Using only the URL to represent a webpage has the advantage of making the computation easier since it does not require fetching the actual content of the webpage associated with the URL. But as a downside, the important information contained in the content does not get utilized. These works are largely orthogonal to our contributions and our pre-training can be applied to their method of utilizing structural information or other richer features from webpages as well. Therefore we limit our work to using the URL and simple content features: title and description.  

\noindent
\textbf{Graph Representations:} Many approaches have been developed that utilize link structure to embed nodes in graphs~\citep{el2022graph}. Shallow approaches such as node2vec~\citep{grover2016node2vec} and TwHIN~\citep{el2022twhin} directly learn an embedding vector for each node. Deeper GNN approaches utilize higher-order interactions among nodes to learn inductive node representations \citep{hamilton2017inductive, ying2018graph}. The key difference between these approaches and ours is that these approaches are content-agnostic and represent each node by its identifier. This makes them transductive and not inductive, i.e. the graph needs to be re-trained in order to obtain representations for new content. For online social networks, this is not scalable given how fast new content gets created and shared and used in downstream tasks such as content recommendation, content classification, etc. As compared to this, our approach is based on webpage content and can be applied on the fly as the content gets generated.


\section{Preliminaries}
\label{sec:prelim}
Let $\mathcal{P} = \{p_1, p_2, \ldots p_N\}$ be a set of Users on a social network and $\mathcal{W} = \{w_1, w_2, \ldots w_M\}$ be the set of webpages as indicated by URLs and $\mathcal{T} = \{t_1, t_2, \ldots t_M\}$ be the set of webpage content associated with each URL.  Let $\mathcal{G}$ constitute a bipartite graph representing the engagements between users ($\mathcal{P}$) and webpage URLs ($\mathcal{W}$).

Based on the engagements in $\mathcal{G}$, we seek to learn for each user, $p_i$ a $d$-dimensional embedding vector $\mathbf{p^g_i} \in \mathbb{R}^d$; similarly for each target item $w_j$ an embedding vector $\mathbf{w^g_j} \in \mathbb{R}^d$. We call these engagement-based embeddings, and assume that they model user-webpage relevance $p(\textrm{relevance} | p_i, w_j) = g(\mathbf{p_i}, \mathbf{w_j})$ for a suitable function $g$. 

Given these engagement-based Webpage embeddings, we seek to take the URL and text content for each webpage, and continue pre-training of the language model on an embedding-alignment pre-training task. We next describe the proposed training process in detail. For the reader's convenience, the main symbols used are captured in~\autoref{tab:symbols}.


\section{Webpage Representation Training}
\label{sec:dec}

\begin{figure}[ht]
\begin{center}
\includegraphics[width=\textwidth]{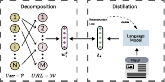}
\end{center}
\caption{Proposed approach consists of two stages: (1) decomposing user URL engagement graph, followed by (2) distilling the learnt URL representations in a language model, incorporated through pre-training the model.}
\label{fig:overview}
\end{figure}

The proposed approach has two stages, with different motivations. This is shown in~\autoref{fig:overview} and described below.

\begin{table}[t]
\caption{List of symbols and notations used in this paper.}
\label{tab:symbols}
\begin{center}
\begin{tabular}{ll}
\multicolumn{1}{c}{\bf Symbol}  &\multicolumn{1}{c}{\bf Description}
\\ \hline \\
        $\mathcal{P}$  & Set of users \\
        $\mathcal{W}$  & Set of webpage URLs \\
        $\mathcal{E}$  & Set of edges \\
        $\mathcal{T}$  & Set of tokenized webpage content \\
        $L$ & Dimensionality of tokenized content \\
        $p$ or $p_i$ & A user $\in \mathcal{P}$ \\
        $w$ or $w_j$  & A webpage URL $\in \mathcal{W}$ \\
        $w^g_j$ & Graph based webpage representation $\in R^d$ \\
        $t_j$ & Content based webpage tokens $\in R^L$ \\
        $t_j$ & Content based webpage representation $\in R^d$ \\
        $t^u_k, t^t_k, t^d_k$ & Tokens from URL, title descriptions \\
\end{tabular}
\end{center}
\end{table}

\begin{enumerate}
    \item \textbf{Decomposition.} 
    The goal of this stage is to decompose the user URL engagement graph and obtain engagement-based URL representation. At this stage we only use URL identifiers;  their content is not used. We first construct a  bipartite graph $\mathcal{G}$ with $\mathcal{P}$, $\mathcal{W}$ as described in~\autoref{sec:prelim}. $\mathcal{E}$ represents the set of edges in $\mathcal{G}$. An edge exists between a user $p \in \mathcal{P}$ and a URL $w \in \mathcal{W}$ if $p$ has ``engaged" with $w$ in the graph training window. The definition of an engagement is up to the user's design. For the purpose of our experiments we have used engagements to mean a union of activities such as a user `Favorite'', ``Reply'', ``Retweet'' or ``Share'' a Tweet containing a URL. Several algorithms are available that can be used to decompose the graph consisting of $\mathcal{P}$, $\mathcal{W}$ and $\mathcal{E}$, for example Node2Vec~\citep{grover2016node2vec} and TwHIN~\citep{el2022twhin}. We follow the approach outlined in~\cite{el2022knn} to obtain the User and Webpage embeddings, i.e., $\mathbf{p^g_i}$ and $\mathbf{w^g_j}$ respectively corresponding to user $p_i$ and url $w_j$. The output of this stage is $\mathbf{w^g_i} \in \mathbb{R}^d$ for each unique webpage URL $w_i$ in the graph training window. Specifically,
    \begin{equation}
        \mathbf{w^g_i} = g(w_i | \theta) \quad \forall i \in \mathcal{W}
    \end{equation}
    where $g()$ represents the graph decomposition algorithm used and $\theta$ captures the learnt parameters of the graph used to represent each node.
    
    \item \textbf{Distillation.} 
    The goal of this stage is to incorporate the learnt representations $\mathbf{w^g_i}$ from the decomposition stage into the language model. 
    We use Transformer architecture~\cite{vaswani2017attention} based language models and we design this stage as continued pre-training of the models. 
    At a high level, the content of a webpage is tokenized and encoded using the language model. We then try to reconstruct the URL representations $\mathbf{w^g_i}$ from the Decomposition stage. Specifically, we use the multilingual BERT base model\footnote{https://huggingface.co/bert-base-multilingual-cased}, mBERT, initialized with pre-trained weights. We use the URL along with the title and description as content from the webpage. 
    The description here is defined as the textual content from the body tag of the webpage HTML. 
    The URL, title, and description are concatenated and tokenized, the output for $w_i$ is $\{[CLS],t^u_{i,1}, t^u_{i,2}, \ldots ,t^t_{i,1},t^t_{i,2}, \ldots ,t^d_{i,1},t^d_{i,2}, \ldots ,[SEP]\}$. 
    Where $t^u, t^t, t^d$ are URL, title and description tokens respectively. We set the maximum token size by looking at the 95th percentile of token length from the training set and set it to 160---i.e., when the tokenized output exceeds 160 we truncate the description. 
    The tokenized output is passed to mBERT to get the embedding corresponding to the $[CLS]$ token, $e_{i, [CLS]}$ which is used to get the webpage representation $l_i \in R^d$ by 
    
    \begin{equation}
    \label{eq:pooled}
    l_i = tanh(W_{pooler}(e_{i, [CLS]}))
    \end{equation}

    Where $W_{pooler}()$ is a fully-connected layer of size 128 with $tanh()$ activation function. For convenience, we re-write~\autoref{eq:pooled} as 
    \begin{equation}
    \label{eq:distill}
    l_i = f(t_i)
    \end{equation}
    
    where $t_i$ $\in R^L$ are the tokens for webpage $w_i$.
    
    The distillation stage tries to increase the cosine similarity between $f(t_i)$ and $g(w_i)$ vs the in-batch negatives. Specifically, following \cite{gao2021simcse} the loss per URL $w_i$ that we optimize at this stage is

    \begin{equation}
    L_i = -log\frac{e^{CosSim(f(t_i), g(w^g_i))/\tau}}{\sum_{j=0}^B{e^{CosSim(f(t_i), g(w^g_j))/\tau}}}
    \end{equation}
    
    where $B$ is the batch size and $CosSim()$ represents cosine similarity. $g()$ and $f()$ are learnable non-linear projections as part of the model learned through the Decomposition and Distillation stages respectively. $\tau$ is the temperature hyperparameter set to 0.01 in our experiments. We train for 3 epochs with a batch size of 128 and a learning rate of $3e-5$.
\end{enumerate}

\section{Experiments}

In order to evaluate the effectiveness of our approach and to quantify its value on downstream personalization and classification applications, we utilize three downstream tasks - (a) Tweet Hashtag prediction, (b) URL topic classification, (c) User-URL engagement prediction. We test URL-BERT under the resource-scare setting of few-shot learning. Under this setting, results are presented for varying numbers of training instances with a fixed test set. \autoref{tab:dataset} provides details about the three tasks. The experimental setup and results are given below. We demonstrate our decomposition and distillation approach using mBERT, and so the latter is used as a baseline. For all downstream tasks, we use URL-BERT and freeze the encoder. This is because in a practical setup, the encoded representation of URLs are attached with each URL in our stack as soon as the URL is shared on Twitter. This facilitates easy reuse by multiple downstream teams. For a fair comparison, the mBERT baseline is similarly frozen. 

\begin{table} 
\caption{Here $\mathcal{|C|}$ is the number of classes in each downstream task along with the test set size.\label{tab:dataset}}
\newcolumntype{C}{>{\centering\arraybackslash}X}
\begin{tabularx}{\textwidth}{lCC}
    \toprule
    \textbf{Dataset} & \textbf{$|\mathcal{C}|$}	& \textbf{Test size}\\
    \midrule
    Tweet Hashtag Prediction		& 43 & 230K\\
    URL Classification		& 15 & 150K\\
    User-URL Engagement		& 2 & 90K\\
    \bottomrule
\end{tabularx}
\end{table}

We train separate classifiers to minimize the cross-entropy loss for each downstream task using~\autoref{eq:classifier}

\begin{equation}
\label{eq:classifier}
x^{'}_i = act(W_{C}(x_i))
\end{equation}

Where $x_i$ is task-dependent input representation, $W_C$ is a fully-connected layer of size 128, $act()$ is non-linearity, we use $tanh()$ for ~\ref{sec:url}, ~\ref{sec:url-eng} and $ReLU()$ for ~\ref{sec:hashtag}. $x^{'}_i$ is passed to a softmax classification layer to get logits. We do not use Tweet text for any of the Twitter internal downstream tasks since the goal is to evaluate the webpage representations of URL-BERT so we do not incorporate any additional features. To stay true to the real-world few-shot setting where a validation set is not available, we have used fixed hyperparameters without any tuning on a held-out set. We use learning rate 1e-5, batch size 8 and epochs 10.

\subsection{Tweet Hashtag prediction}
\label{sec:hashtag}

Using the URL contained in a given Tweet, the goal of this evaluation task is to predict the hashtags that the author may have used in the given Tweet. Our hypothesis is that our proposed approach to pre-train the BERT language model succeeds in capturing user engagement information to represent a given URL, and this in turn leads to improved performance at this task.

This task is evaluated on test dataset containing over 230K Tweets that were created between Sep 7, 2022 to Oct 15, 2022. This test dataset contains 43 hashtags in a multi-class setting. When a Tweet contained more than one hashtag, the one that appeared in more Tweets was kept. 

For this task, $x_i$ in~\autoref{eq:classifier} is the mean pooling over tokens from URL, title and description. The model is trained on the Tweets sampled from Jun 1, 2022 to Aug 31, 2022 containing the same 43 hashtags as in the test dataset.
    
\begin{table} 
\caption{URL-BERT clearly outperforms mBERT baseline
for both URL classification and hashtag prediction task. Samples here represents the number of examples per class that are used for fine-tuning of each model.\label{tab:dmoz_hashtag_table}}
\newcolumntype{C}{>{\centering\arraybackslash}X}
\begin{tabularx}{\textwidth}{CCCCCC}
\toprule
    \multirow{2}{*}[-4pt]{\textbf{Task}} &
    \multirow{2}{*}[-4pt]{\textbf{Samples}} &
    \multicolumn{2}{c}{\textbf{Macro F1}} &
    \multicolumn{2}{c}{\textbf{Micro F1}}\\ 
    \cmidrule(lr){3-4}
    \cmidrule(lr){5-6}
    && mBERT & URL-BERT & mBERT & URL-BERT \\
    \midrule
    \multirow{2}{*}[-1pt]{\thead{{\parbox{1.8cm}{\centering URL topic classification}}}} &8 & 1.03 & 1.64 & 7.03 & 7.41\\
    &64 & 0.97 & 2.70 & 7.025 & 8.37\\
    &512 & 11.23 & 42.00 & 16.60 & 46.64\\
    \midrule[0.01pt]
    \multirow{7}{*}[-1pt]{\thead{{\parbox{1.8cm}{\centering Tweet Hashtag Prediction}}}} &8 & 0.75 & 4.69 & 2.05 & 10.41\\
    &16 & 1.71 & 3.68 & 5.37 & 9.17\\
    &64 & 7.61 & 12.62 & 17.27 & 25.74\\
    &128 & 13.84 & 20.57 & 26.15 & 38.55\\
    &256 & 26.57 & 35.34 & 47.00 & 61.01\\
    &512 & 30.12 & 40.01 & 52.83 & 66.31\\
    &1,000 & 37.38 & 45.29 & 61.95 & 70.20\\
\bottomrule
\end{tabularx}
\end{table}

\autoref{tab:dmoz_hashtag_table} shows the Micro and Macro averaged F1 scores of the baseline as compared to the proposed approach. It can be seen that our proposed approach consistently performs better than the baseline mBERT model. The improvement in performance is more prominent when fewer samples are available to train the classification layer. This shows that the proposed approach is able to improve the URL representation much better than the baseline.

\subsection{URL topic classification}
\label{sec:url}

Classifying a URL refers to assigning a topic, from a pool of pre-defined topic labels, to the webpage that a URL points to. We use DMOZ\footnote{\url{https://dmoz-odp.org/}}, formerly known as Open Directory Project (ODP). DMOZ is a popular web directory with URL, title and description for websites and their corresponding categories. There are 15 such high-level categories that we restrict our experiments to. Since we use a multi-lingual backbone model, we do not drop any non-English examples. We randomly sampled \textit{N} samples per class for fine-tuning the model. From the remaining instances of each class, we sampled 10,000 examples per class for the test set, unless that class had less than 10,000 examples, in which case we retained the maximum number of examples from that class. This results in around 150K test set size. In~\autoref{eq:classifier}, $l_i$ from~\autoref{eq:pooled} is used as $x_i$.~\autoref{tab:dmoz_hashtag_table} shows the performance of our approach when compared to the baseline.

\subsection{User-URL engagement prediction}
\label{sec:url-eng}
For this task, given a user and a Tweet containing a URL, the task is to predict whether the user will \textit{engage} with the Tweet. For this task, our fine-tuning dataset consisted of Tweets from Sept 5, 2022 to Sept 8, 2022 and the test dataset consisted of Tweets from  Sept 15, 2022 to Sept 16, 2022. All Tweets were downsampled by 10\%. Only those URLs were selected that were authored by at least 5 users, i.e., each distinct webpage was shared by at least 5 unique users. The test dataset consisted of over 52K negatives and over 38K positives. Positive label here means that the user ``Favorited'', ``Shared'', ``Retweeted'' a Tweet containing the corresponding URL. A negative label here means that the user saw a Tweet containing the corresponding URL but did not engage via any of the above actions. To obtain this data we downsampled all Tweets containing URLs to 5\% and then downsampled negatives to 4\% to balance the priors in the dataset.

\begin{equation}
\label{eq:url-eng-rep}
x_i = \mathbf{p^{g}_i} \oplus l_i
\end{equation}

\begin{table} 
\caption{PR-AUC for the User-URL fav task on a test set size of around 90k. URL-BERT consistently performs better than the baseline, especially as number of samples increases.\label{tab:URL-fav}}
\newcolumntype{C}{>{\centering\arraybackslash}X}
\begin{tabularx}{\textwidth}{CCC}
\toprule
\textbf{Samples}	& \textbf{mBERT}	& \textbf{URL-BERT}\\
\midrule
8 & 0.415 & 0.417\\
16 & 0.416 & 0.421\\
64 & 0.417 & 0.431\\
128 & 0.427 & 0.462\\
256 & 0.433 & 0.499\\
512 & 0.457 & 0.562\\
1,000 & 0.521 & 0.633\\
\bottomrule
\end{tabularx}
\end{table}

As it can be seen through \autoref{tab:dmoz_hashtag_table} and \autoref{tab:URL-fav}, our approach clearly outperforms the baseline. These experiments demonstrate that URL-BERT is able to utilize user engagement information to better represent the URLs such that this improves the performance on multiple diverse downstream tasks. After the continued pre-training to align content-based URL representations to engagement-based ones, URL-BERT is better able to capture the nuances of how users interact with webpages solely from the content.

\section{Conclusion}

Pre-trained language models have successfully been used for a plethora of downstream NLP tasks. But fine-tuning them for a particular task requires access to a large annotated dataset. In this work, we have presented a pre-training task to integrate user interaction with URLs into pre-trained LMs. The method we present is independent of the features used to represent the webpage or a particular language model. This results in the LM having inherent information on how to better represent a webpage using only the semantic content but still instilling user information. Consequently, a new downstream task involving a webpage only requires a few examples to get better results than an off-the-shelf model.


\bibliographystyle{iclr2024_conference}

\end{document}